\documentclass[10pt,twocolumn,letterpaper]{article}

\usepackage{iccv}
\usepackage{times}
\usepackage{epsfig}
\usepackage{graphicx}
\usepackage{amsmath}
\usepackage{amssymb}
\usepackage{comment}
\usepackage{authblk}
\usepackage{threeparttable}
\usepackage[breaklinks=true,bookmarks=false]{hyperref}

\iccvfinalcopy
\ificcvfinal\fi
\def\iccvPaperID{10}

\begin{document}

\specialcomment{acks}{%
  \begingroup
  \section*{Acknowledgments}
  \phantomsection\addcontentsline{toc}{section}{Acknowledgments}
}{%
  \endgroup
}

\title{Leveraging Weakly Annotated Data for Fashion Image Retrieval and Label Prediction \vspace{-2ex}}

\author[1]{Charles Corbi\`{e}re}
\author[1,2]{Hedi Ben-Younes}
\author[1]{Alexandre Ram\'{e}}
\author[1]{Charles Ollion}
\affil[1]{Heuritech, Paris, France}
\affil[2]{UPMC-LIP6, Paris, France}
\affil[ ]{{\tt\small \{corbiere,hbenyounes,rame,ollion\}@heuritech.com}}

\maketitle
\thispagestyle{empty}

\begin{abstract}
   In this paper, we present a method to learn a visual representation adapted for e-commerce products. Based on weakly supervised learning, our model learns from noisy datasets crawled on e-commerce website catalogs and does not require any manual labeling. We show that our representation can be used for downward classification tasks over clothing categories with different levels of granularity. We also demonstrate that the learnt representation is suitable for image retrieval. We achieve nearly state-of-art results on the DeepFashion In-Shop Clothes Retrieval and Categories Attributes Prediction \cite{10.1109/CVPR.2016.124} tasks, \emph{without} using the provided training set. 
\end{abstract}

\vspace{-0.4cm}
\section{Introduction}
While online shopping has been an exponentially growing market for the last two decades, finding exactly what you want from online shops is still not a solved problem.
Traditional fashion search engines allow consumers to look for products based on well chosen keywords. Such engines match those textual queries with meta-data of products, such as a title, a description or a set of tags.
In online luxury fashion for instance, they still play an important role to address this customer pain point: 46\% of customers use a search engine to find a specific product; 31\% use it to find the brand they're looking for \footnote{http://www.mckinsey.com/business-functions/marketing-and-sales/our-insights/the-opportunity-in-online-luxury-fashion}.
However, those meta-data informations may be incomplete, or use a biased vocabulary. For instance, a description may denote as "marini\`ere" a long sleeves shirt with blue/white stripes. It then appears crucial for online retailers to have a rich labeled catalog to ensure good search. Moreover, these search engines don't incorporate the visual information of the image associated to the product.

\begin{figure}[h]
\includegraphics[width=6.75cm]{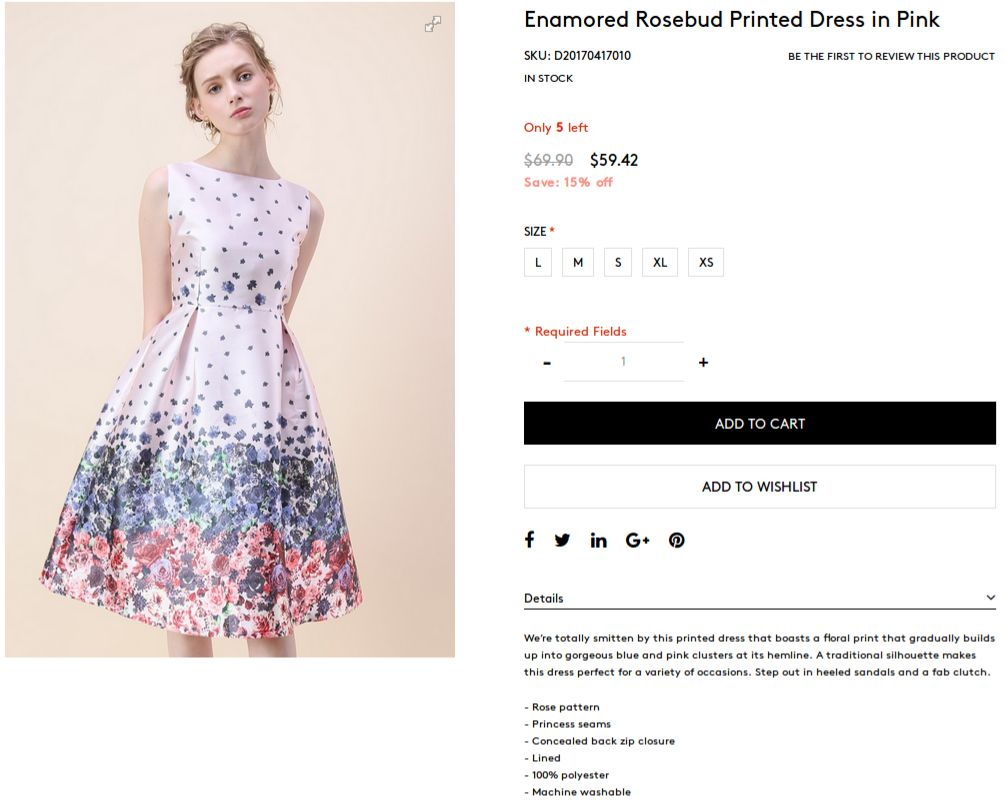}
\caption{\label{fig:sample} Our dataset is composed of images and a few associated text descriptors such as their title and their	 description}
\end{figure}

Computer vision for fashion e-commerce images has drawn an increasing interest in the last decade. It has been used for similarity search \cite{conf/cvpr/WangSLRWPCW14, conf/iccv/KiapourHLBB15, conf/mir/WangSZZJ16,DBLP:journals/corr/ShankarNAKC17}, automatic image tagging \cite{Kalantidis:2013:GLC:2461466.2461485,Bossard:2012:ACS:2482108.2482136}, fine-grained classification \cite{470,10.1109/CVPR.2016.124} or N-shot learning \cite{conf/nips/BergamoT10}. In all of these tasks, a model's performance is highly dependent on a visual feature extractor. Using a Convolutional Neural Network (CNN) trained on ImageNet \cite{Deng09imagenet:a} provides a good baseline. However, there are two main problems with this representation. First, it has been trained on an image distribution that is very far from e-commerce, as it has never (or rarely) seen such pictures. Second, the set of classes it has been trained on is different from a set of classes that could be meaningful in e-commerce. A useful representation should separate different types of clothing (\textit{e.g.} a skirt and a dress), but it should also discriminate between different lengths of sleeves for shirts, trouser cuts, types of handbags, textures, colors, shapes,... 

\begin{figure*}[t]
\begin{center}
\includegraphics[width=14.25cm]{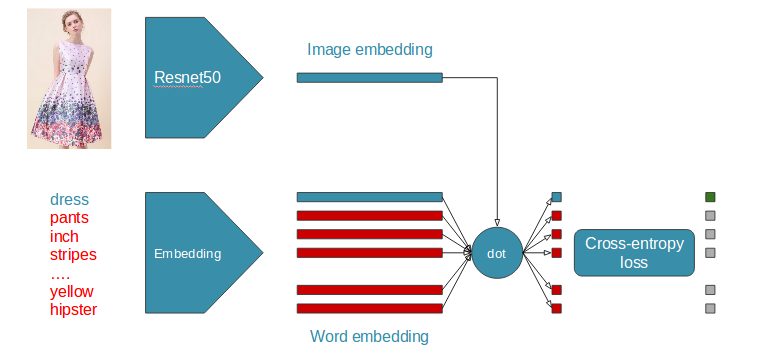}
\caption{\label{fig:model} Training of our model: predict one label, picked from the bag-of-words description, from an image. Both image and words are embedded before being coupled in a dot product. We output a probability for each word in the vocabulary. }
\end{center}
\end{figure*}

Our goal is to learn a visual feature extractor designed for e-commerce images. This representation should:
\begin{itemize}
\vspace{-0.35cm}
\item encode multiple levels of visual semantics: from low level signals (color, shapes, textures, fabric,...) to high level information (style, brand),
\vspace{-0.35cm}
\item be separable over visual concepts, so we can train very simple classifiers over clothing types, colors, attributes, textures, \textit{etc.}, 
\vspace{-0.35cm}
\item provide a meaningful similarity between images, so we can use it in the context of image retrieval.
\end{itemize}

To these ends, we train a visual feature extractor on a large set of weakly annotated images crawled from the Internet. These annotations correspond to the textual description associated to the image. The model is learned on a dataset at \emph{zero} labeling cost, and is exclusively constituted of data points extracted from e-commerce websites. Our main contribution is an in-depth analysis of the model presented in \cite{DBLP:journals/corr/JoulinMJV15}, through applications to fashion image recognition tasks such as image retrieval and attribute tagging. We also improved the method by upgrading the CNN architecture and dealt with multiple languages, mainly English and French.
In Section~\ref{section:model}, we explain the model, how we handle noise in the dataset, as well as some implementation details. In Section~\ref{section:experiments}, we provide results given by our representation on image retrieval and classification, over multiple datasets. Finally in Section~\ref{section:future}, we conclude and go over some possible improvement tracks.

\section{Learning Image and Text Embeddings with Weak Supervision}\label{section:model}
One major issue in applied machine learning for fashion is the lack of large clothing e-commerce datasets with a rich, unique and clean labeling. Some very interesting work has been done on collecting datasets for fashion \cite{conf/iccv/KiapourHLBB15,10.1109/CVPR.2016.124}. However, we believe it is very hard to be exhaustive in describing every visual attribute (pieces of clothing, texture, color, shape, \textit{etc.}) in an image. Moreover, even if this labeling work could be perfectly carried, it would come at very high cost, and should be manually done each time we wanted to add a new attribute. A possible source of annotated data is the e-commerce website catalogs. They provide a great amount of product images associated with descriptions, such as the one in Figure~\ref{fig:sample}. While this description contains information about the visual concepts in the image, it also comes with a lot of noise that could harm the learning. 

We explain now the approach we used to train a visual feature extractor on noisy weakly annotated data.
\subsection{Weakly Supervised Approach}
Learning with noisy labeled training data is not new to the machine learning and computer vision community \cite{5be119761e7f4d1080d03ba826c2301e,NIPS2013_5073,DBLP:journals/corr/SukhbaatarF14}. Label noise in image classification \cite{conf/cvpr/XiaoXYHW15} usually refers to errors in labeling, or to cases where the image does not belong to any of the classes, but mistakenly has one of the corresponding labels. In our setting, in addition to these types of noise, there are some labels in the classes vocabulary that are not relevant to any input. Text descriptions are noisy as they contain common words (\textit{e.g.} 'we', 'present'), subjective words (\textit{e.g.} 'wonderful') or non visual words (\textit{e.g.} 'xl', 'cm'), which are not related to the input image. As we don't have any prior information on which labels are relevant and which are not, we keep the preprocessing of textual data as light as possible.

\begin{figure*}[t]
\begin{center}
\includegraphics[height=5cm, width=15cm]{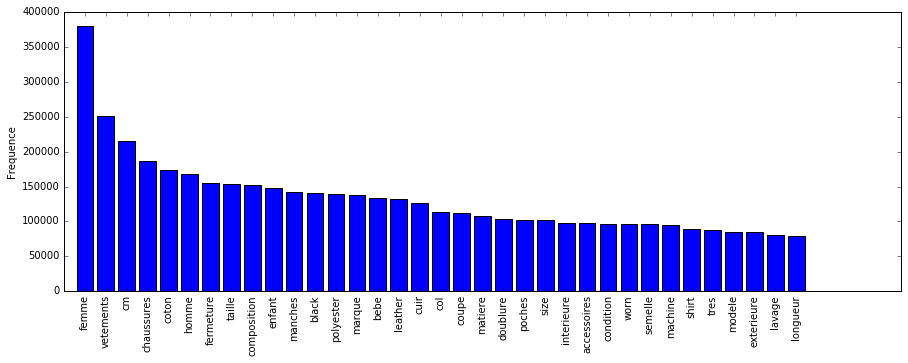}
\caption{\label{fig:distrib} Most frequent labels in training dataset.}
\end{center}
\end{figure*}

\subsection{Model}
Our work builds upon the one presented in \cite{DBLP:journals/corr/JoulinMJV15}, which we explain in this section. The model's training scheme is exposed in Figure~\ref{fig:model}

Let $x \in \mathcal{I}$ be an image, and $y \in \{0,1\}^K$ the associated multi-label vector, such that $\forall k \in [1, K], y^k = 1$ if the k-th label of the vocabulary is true for the image $x$. 
We use a CNN to compute a visual feature $z = f(x, \theta) \in \mathbb{R}^I$, where $\theta$ are the weights of our convolutional neural network.
This image embedding is given to a classification layer:
\begin{equation}
\hat{y} = softmax\left( W^T z \right)
\end{equation}
where $W \in \mathbb{R}^{I \times K}$. Note that for all $k \in [1,K]$, the column vector in $w_k = W[:,k]$ corresponds to the embedding of the k-th word in the vocabulary.

\begin{figure*}[t]
\includegraphics[width=17cm]{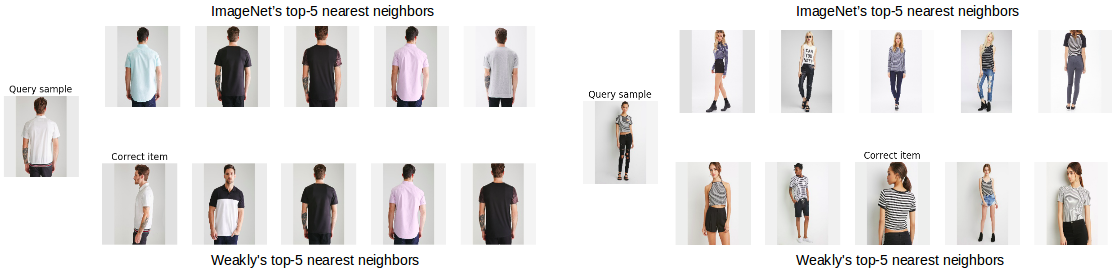}
\caption{\label{fig:most_sim} Comparison between visual similar images from DeepFashion In-Shop dataset according to our weakly learned visual features and ImageNet based visual features. Our representation seems more robust to human pose (on the left) and successfully captured fine-grained concepts such as stripes (on the right).}
\end{figure*}

\subsection{Label imbalance management}
\label{subsection:imbalance}
As seen on Figure~\ref{fig:distrib}, the distribution of words in our dataset is highly unbalanced. 
Due to our minimal preprocessing, we observe a high frequency for some non-visual words such as "xl", "cm" or "size" as they appear very frequently in descriptions. Many examples contain those non-visual words that our model would be asked to predict, which is likely to harm the training.

To overcome this issue, and as it was done in \cite{DBLP:journals/corr/JoulinMJV15}, we perform {\itshape uniform} sampling. Specifically, during training, we sample uniformly a word $w$ from the vocabulary. We then randomly choose an image $x$ whose bag-of-words contains $w$ and we try to predict $w$ given $x$.
\subsection{Loss}
As we want to predict one label among a vocabulary $K$ for each image, we use the cross-entropy loss. It minimizes the negative sum of the log-probabilities over all positive labels 

\begin{equation}
 L(\theta, W, \mathcal{D}) = -\frac{1}{N} \sum_{n=1}^{N} \sum_{k=1}^K y^k_{n} \log \frac{\exp{(w_k^T f(x_n,\theta))}}{\sum_{i=1}^K \exp{(w_i^T f(x_n,\theta))}}
\end{equation}

\subsection{Implementation details}
\subsubsection*{Negative sampling} \vspace{-0.25cm} We operate in a context where the vocabulary can be of arbitrary size. Computing probabilities for all those classes for each sample can be very slow. Negative sampling \cite{NIPS2013_5021} is one way of addressing this problem. After selecting a positive label for an image sample, we randomly draw $N_{neg}$ negative words within the vocabulary. We compute the scores and the softmax only over those chosen words. 

\vspace{-0.25cm}

\subsubsection*{Learning}  \vspace{-0.25cm}
We trained our model with stochastic gradient descent (SGD) on batch of size 20. We consider that an epoch is achieved when the model saw a number of images equivalent to $1/10$ of the dataset size, which is approximately 1.3M images in total. After each epoch, we compute a validation error based on a held-out validation set. The initial rate was set to 0.1 and divided by 10 after 10 epochs without improvement. We stop the training after 20 epochs without improvements on our validation dataset. We use the ResNet50 architecture \cite{He2015} for the visual feature extractor $f(x,\theta)$, with pre-trained weights on ImageNet. Because the last layer has been initialized randomly, we start by learning only the last layer $W$ for 20 epochs, and then we fine-tune the parameters $\theta$ in the CNN. 

\subsection{Training dataset}
We built a dataset of about 1,3M images and their associated labels from multiple e-commerce website sources, mostly French, English and Italian. We crawled most of the time one image per product, except when multiple images where available. In that case, we consider them as four different samples with same associated bag of labels.

For each source, we select the relevant fields to keep (title, category name, description,...) and concatenate them. After lower-casing and removing punctuation, we use the RegexpTokenizer provided by NLTK \cite{Loper02nltk} to get a list of words. We remove stopwords, frequent non-relevant words (name of the website, 'collection', 'buy',...) and non alphabetic words. Our final dataset is a list of product images, associated to their respective bag-of-words obtained by the previous preprocessing.

We have deliberately kept preprocessing as minimal as possible, so it is easy to scale to many sources. Thus, we need our model to adapt to this noise in the data. After preprocessing and aggregating the multiple sources, our final vocabulary is composed of 218,536 words. We chose to restrain the vocabulary to the 30,000 most frequent words. The average number of labels per sample is 26,88. We split our dataset into a training and a validation dataset. The validation set is made of the same labels as the training dataset and represent 0.5\% of the total size. \\

\vspace{-0.4cm}

\section{Experiments and evaluation} \label{section:experiments}
After learning the representation on our large weakly annotated dataset, we want to evaluate this representation. To what extent is this representation useful for tasks such as garment classification, attribute tagging or image retrieval ?

\subsection{Evaluation datasets}
\vspace{-0.1cm}

We evaluate our representation on 5 datasets: two public datasets (DeepFashion) used for tagging and image retrieval; three in-house datasets used respectively for category classification, fine-grained classification and image retrieval.

\vspace{-0.3cm}

\paragraph{DeepFashion Categories and Attributes Prediction} 
evaluates the performance on clothing category classification, and on attribute prediction (multi-labelling). It contains 289,222 images for 50 clothing categories and 1,000 clothing attributes. While an image can only be affected to one class, it can be associated to multiple labels. The average number of labels for an image is 3.38. For each image in train and test sets, we select a crop available from a ground truth bounding box.

\vspace{-0.25cm}

\paragraph{DeepFashion In-Shop Retrieval} 
contains 7,982 clothing items with 52,712 images. 3,997 classes are for training (25,882 images) and 3,985 items are for testing (28,760 images). The  test  set  is composed of a  query  set  and a gallery set, where query set contains 14,218 images of 3,985 items and database set contains 12,612 images of 3,985 items. As in the Categories and Attributes Prediction benchmark, we cropped each image using a ground truth bounding box.

\vspace{-0.25cm}

\paragraph{ClothingType}
We have labeled a dataset with 18 classes, each one corresponding to a garment type (\textit{e.g.} bag, dress, pants, shoes, ...). This in-house dataset contains approximately 736,000 images.

\vspace{-0.25cm}

\paragraph{HandBag}
In addition to the previous dataset, this in-house dataset focuses on bags for fine-grained recognition. Here, the differences between classes are more subtle: bucket bag, doctor bag, duffel bag, etc... It contains 3,060 samples within 13 classes, each one corresponding to a specific type of handbag.

\vspace{-0.25cm}

\paragraph{Dress Retrieval}
This in-house similarity dataset was gathered by crawling an e-commerce website. We collected a list of sets of images, each corresponding to the same item. We used an image classifier to filter out all non-dress items. The final dataset contains 9,009 items for training (20,200 images) and 1,001 items for testing. 
On this dataset, we keep only images where clothing are worn on humans.

\begin{table*}[t]
  \label{tab:deepclassif}
  \resizebox{\textwidth}{!}{%
  \begin{tabular}{lrr|rrrrrrrrrr|rr}
    \hline
    &\multicolumn{2}{c}{Category}&\multicolumn{2}{c}{Texture}&\multicolumn{2}{c}{Fabric}&\multicolumn{2}{c}{Shape}&\multicolumn{2}{c}{Part}&\multicolumn{2}{c}{Style}&\multicolumn{2}{c}{All}\\
    \hline
    &top-3&top-5&top-3&top-5&top-3&top-5&top-3&top-5&top-3&top-5&top-3&top-5&top-3&top-5\\
    \hline
    WTBI \cite{conf/eccv/ChenGG12} & 43.73 & 66.26 & 24.21 & 32.65 & 25.38 & 36.06 & 23.39 & 31.26 & 26.31 & 33.24 & 49.85 & 58.68 & 27.46 & 35.37\\
    DARN \cite{DBLP:journals/corr/HuangFCY15} & 59.48 & 79.58 & 36.15 & 48.15 & 36.64 & 48.52 & 35.89 & 46.93 & 39.17 & 50.14 & 66.11 & 71.36 & 42.35 & 51.95\\
    FashionNet \cite{10.1109/CVPR.2016.124} & 82.58 & 90.17 & 37.46 & 49.52 & \bf{39.30} & \bf{49.84} & 39.47 & 48.59 & \bf{44.13} & \bf{54.02} & \bf{66.43} & \bf{73.16} & \bf{45.52} & \bf{54.61}\\
    Lu et al.* \cite{DBLP:journals/corr/LuKZCJF16} & \bf{86.72} & 92.51 & - & - & - & - & - & - & - & - & - & - & - & -\\
    Weakly& 86.30 & \bf{92.80} & \bf{53.60} & \bf{63.20} & 39.10 & 48.80 & \bf{50.10} & \bf{59.50} & 38.80 & 48.90 & 30.50 & 38.30 & 23.10 & 30.40 \\
\end{tabular}
}

\begin{tablenotes}
      \footnotesize
      \item *Attribute scores not tested.
\end{tablenotes}
  \caption{Performance of category classification and attribute prediction on DeepFashion dataset}
\end{table*}

\subsection{Image retrieval}
\vspace{-0.1cm}
In this task, given a query image containing an item, we aim at retrieving images that contain the same item. To do so, we compute the score between two images using the cosine similarity between their representation. For a given query image, we sort all gallery images in decreasing order of similarity, and evaluate our retrieval performance using top-k retrieval accuracy, as in \cite{10.1109/CVPR.2016.124, DBLP:journals/corr/YuanYZ16}. For a given test query image, we give the model a score of 1 if an image of the same item is within the k highest scoring gallery images, 0 else. We adopt this metric for both our image retrieval datasets (DeepFashion In-Shop Retrieval and Dress Retrieval).

\begin{figure}[h]
\begin{center}
\includegraphics[width=8.5cm]{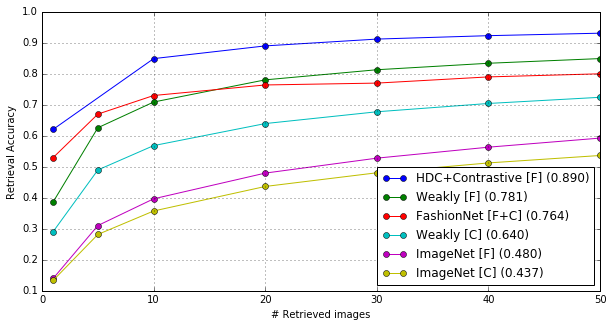}
\caption{\label{fig:scores_similarity} Retrieval accuracy for top-k (k=1,5,10,20,30,40,50). We give the top-20 retrieval accuracy between brackets for each model in the caption.}
\end{center}
\end{figure}

In Figure~\ref{fig:scores_similarity}, we show the results on top-k retrieval accuracy on DeepFashion In-shop Retrieval dataset, for multiple values of k. FashionNet corresponds to the model presented in \cite{10.1109/CVPR.2016.124}, and HDC+Contrastive is the model in \cite{DBLP:journals/corr/YuanYZ16}. We denote by [F] (resp. [C]) models that use the full image (resp. an image cropped on the product) to compute retrieval scores. We provide the ImageNet baseline both [F] and [C] models, where we use as feature extractor the penultimate layer of a CNN trained on ImageNet. We would like to emphasize on the fact that our Weakly model, as well as the ImageNet baseline, \emph{do not use the training set} of DeepFashion In-shop Retrieval, unlike HDC+Contrastive and FashionNet. 

\vspace{-0.5cm}

\begin{figure*}[t]
\begin{center}
\includegraphics[width=11cm]{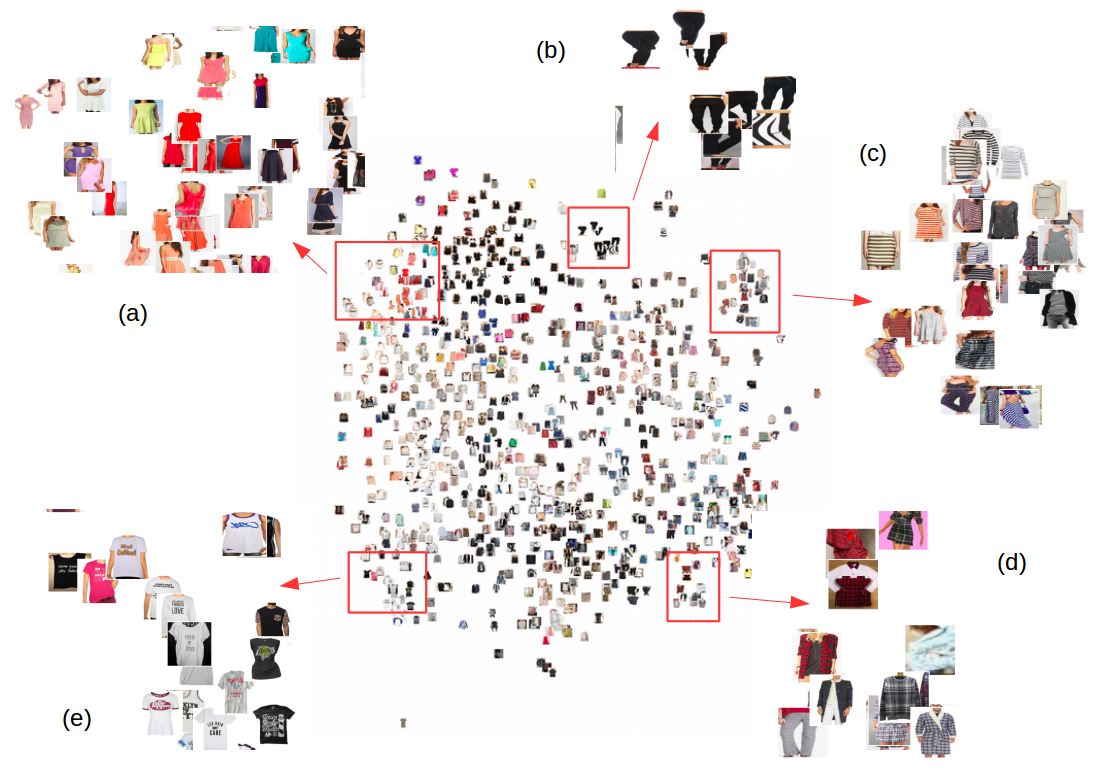}
\caption{\label{fig:tsne} t-SNE map of 1,000 image samples from DeepFashion Categories dataset based on our Weakly image features extractor. We can identify some local subcategories, such as colorful dresses (a), black pants (b), stripes (c), checked (d) or printed shirt (e).}
\end{center}
\end{figure*}

First, we note that not using bounding boxes for our ImageNet baseline or our Weakly model considerably increase accuracy. Our intuition is that human models in the DeepFashion In-shop dataset often wear the same ensemble of items together, meaning for one shirt item considered for instance, the human model would be wearing the same pants and shoes on all item's image. As a consequence of this bias, it seems easier to evaluate similarity on a ensemble of clothings than on a single clothing on this dataset. 

Our Weakly model without crop performs as well as FashionNet, and even outperforms it when $k \geq 20$: considering top-20 retrieval accuracy, it predicts the correct item 78,1\% of the time, against 76,4\% for FashionNet. Besides, in both the [F] and [C] setups, our Weakly model improves over the ImageNet baseline (from 48\% to 78.1\% for [F], and from 43.7\% to 64.0\% for [C]). This validates our hypothesis that our model has learned a specific e-commerce representation. In Figure~\ref{fig:most_sim}, we show an example of a query image, its top-5 similar images according to our weakly learned visual features, and its top-5 similar images according to ImageNet based visual features. As we can see, the similarity encoded by network trained on ImageNet brings together products that are on a same coarse semantic concept, while our representation encodes a more precise and rich closeness, which is based not only the image type, but also on their shape, texture, and fabric. Plus, our representation seems less dependent to human model's pose. 

On our in-house dress retrieval dataset, we also observed that the Weakly model improved over ImageNet Model on retrieval accuracy. The Weakly model obtained a top-20 retrieval accuracy of 83,71\%, against 65,65\% for the ImageNet model. Once again, we point out that we do not perform any training on the retrieval task of this dataset.

\subsection{Tagging}
\vspace{-0.1cm}
\begin{table*}[t]
  \label{tab:classif}
  \resizebox{\textwidth}{!}{%
  \begin{tabular}{lrrrrrrrrrrrrrrrrrr}
    \hline
    &bag & belt & body & bra & coat & combi & dress & eyewear & gloves & hat & neckwear & pants & shoes & shorts & skirt & socks & top & underpants\\
    \hline
    ImageNet&99.63 & 98.02 & 98.20 & 99.27 & 99.00 & 96.01 & 98.68 & 99.93 & \bf{99.99} & 99.45 & 99.18 & 99.46 & 99.87 & 99.23 & 98.67 & 99.35 & 98.67 & 99.5\\
    Weakly & \bf{99.86} & \bf{99.46} & \bf{99.08} & \bf{99.67} & \bf{99.31} & \bf{98.11} & \bf{99.13} & \bf{99.99} & 99.97 & \bf{99.73} & \bf{99.33} & \bf{99.56} & \bf{99.93} & \bf{99.32} & \bf{99.21} & \bf{99.83} & \bf{99.26} & \bf{99.67}\\
    \hline
\end{tabular}
}
\\
  \caption{AUC classification score for clothing categories}
\end{table*}

\begin{table*}[!htb]
  \label{tab:bags}
  \resizebox{\textwidth}{!}{%
  \begin{tabular}{lrrrrrrrrrrrrrrrrrr}
    \hline
    &backpack & baguette & bowling bag & bucket bag & doctor bag & duffel bag & hobo bag & luggage & clutch & saddle bag & satchel & tote & trapeze\\
    \hline
    ImageNet&95.15 & 87.63 & 90.42 & 94.35 & 90.99 & 87.97 & 92.73 & \bf{87.65} & 96.52 & 91.77 & 88.58 & 96.77 & 92.11\\
    Weakly & \bf{95.94} & \bf{91.85} & \bf{92.13} & \bf{94.87} & \bf{91.59} & \bf{90.12} & \bf{95.19} & 86.96 & \bf{97.24} & \bf{93.12} & \bf{91.45} & \bf{97.64} & \bf{93.61}\\
\end{tabular}
}
\\
  \caption{AUC classification score for fine-grained type of bags}
\end{table*}

We conducted multi-class classification and multi-labelling experiments to assess the quality of our visual representation on transfer learning. On the public DeepFashion Categories dataset, we pre-computed images representation using our Weakly image feature extractor on image crops. Then, we train a simple classifier using a fully-connected layer followed by a softmax activation function. The results are shown on the Table~\ref{tab:deepclassif}, at the column Category. With this simple classifier, our results are on par with the state-of-art model by Lu et al. \cite{DBLP:journals/corr/LuKZCJF16}. 

On DeepFashion Attributes, we train a fully-connected layer with a sigmoid output and a binary cross-entropy loss. As we can see in Table~\ref{tab:deepclassif}, our model significantly improves over previous state-of-the-art on textures and shape labels top-k recall. However, part and style attributes seem more difficult to separate for our Weakly representation. This might be due to the fact that texture-like and shape-like labels are more represented than part and style words in the large weak dataset. This would require further investigation.

We carried out experiments on our in-house ClothingType dataset where images are annotated according to their clothing category (such as bags, shirt, dress, shoes, \textit{etc.}). Table~\ref{tab:classif} shows the improvement on AUC scores over the ImageNet model for each of the clothing categories using our new representation. This indicates that our training scheme was able to learn discriminative features for garment classification. 

Finally, we now focus on a fine-grained recognition task. The HandBag dataset contains images annotated with their specific type of bag. In this dataset, the differences between classes are more subtle than in the ClothingType dataset. The training and evaluation are the same as for the previous experiment. As in the previous experiment, we improved AUC scores for nearly each type of bags (see Table~\ref{tab:bags}). 

\subsection{Exploratory visualization using t-SNE}
\vspace{-0.1cm}
To obtain some insight about our Weakly representation, we applied t-SNE \cite{maaten2008visualizing} on features extracted using our Weakly feature extractor. We did this for 1,000 images from DeepFashion Categories test set. Figure~\ref{fig:tsne} shows full map and some interesting close-ups. On top left (a), we can see a cluster of dresses sub-divised into multiple sub-clusters corresponding to different colors. The cluster (b) shows a focus on black pants. In the zone (c), we can easily see that the model gathered images containing stripes, and it seems like it has separated tops from dresses inside this cluster (with large striped sweaters on top). Checked clothings are grouped in cluster (d), while printed t-shirts are represented in cluster (e). This plot shows that our representation is able to group together concepts that are close in terms of clothing type, texture, color and style.
z
\vspace{-0.1cm}
\section{Discussion and Future Work}\label{section:future}
\vspace{-0.1cm}
We presented in the future a method to learn a visual representation adapted to fashion. This method has the major advantage to overcome the issue of finding a large and clean e-commerce dataset. The results shows clear improvements compared to a visual representation trained on ImageNet, improving performance on multiple tasks such as image retrieval, classification and fine-grained recognition. 

In the future, we would like to investigate on the possibility to better train our visual feature extractor using an external knowledge base of textual concepts.

\begin{acks}
  The authors would like to thank all the Heuritech team for providing an efficient network infrastructure. We'd also like to thank Alexandre Ram\'e for his help on the transfer learning evaluation task.
\end{acks}

{\small
\bibliographystyle{ieee}
\bibliography{egbib}
}
\end{document}